\documentclass{article}

\usepackage{style}

\usepackage[utf8]{inputenc}
\usepackage[T1]{fontenc}
\usepackage{hyperref}
\usepackage{url}
\usepackage{booktabs}
\usepackage{amsfonts}
\usepackage{nicefrac}
\usepackage{microtype}
\usepackage{lipsum}
\usepackage{fancyhdr}
\usepackage{graphicx}
\usepackage{multirow}
\usepackage{rotating}
\usepackage[ruled]{algorithm2e}
\usepackage{algpseudocode}
\usepackage{amsmath}
\usepackage{tablefootnote}
\usepackage{appendix}

\usepackage{xcolor}
\usepackage{listings}
\usepackage{tabularx}
\usepackage{graphicx}
\usepackage{float}
\makeatletter
\renewcommand{\thealgocf}{A\@arabic\c@algocf}
\renewcommand{\fnum@algocf}{\AlCapSty{\AlCapFnt\thealgocf\nobreakspace\algorithmcfname}}%
\makeatother

\title{RISC-V RVV efficiency for ANN algorithms}

\author{
  Konstantin Rumyantsev \\
  YADRO \\
  Nizhny Novgorod\\
  \texttt{k.rumyantsev@yadro.com} \\
   \And
  Pavel Yakovlev \\
  YADRO \\
  Nizhny Novgorod\\
  \texttt{p.yakovlev@yadro.com} \\
   \And
  Andrey Gorshkov \\
  YADRO \\
  Nizhny Novgorod\\
  \texttt{a.gorshkov@yadro.com} \\
   \And
  Andrey P. Sokolov \\
  YADRO \\
  Moscow State University \\
  Moscow\\
  \texttt{andrey.sokolov@yadro.com} \\
}

\begin{document}
\maketitle

\begin{abstract}
Handling vast amounts of data is crucial in today's world. The growth of high-performance computing has created a need for parallelization, particularly in the area of machine learning algorithms such as ANN (Approximate Nearest Neighbors). To improve the speed of these algorithms, it is important to optimize them for specific processor architectures. RISC-V (Reduced Instruction Set Computer Five) is one of the modern processor architectures, which features a vector instruction set called RVV (RISC-V Vector Extension). In machine learning algorithms, vector extensions are widely utilized to improve the processing of voluminous data. This study examines the effectiveness of applying RVV to commonly used ANN algorithms. The algorithms were adapted for RISC-V and optimized using RVV after identifying the primary bottlenecks. Additionally, we developed a theoretical model of a parameterized vector block and identified the best on average configuration that demonstrates the highest theoretical performance of the studied ANN algorithms when the other CPU parameters are fixed.
\end{abstract}


\section{Introduction}

\setlength{\algomargin}{1em}
\SetKw{KwBy}{by}
\SetKwComment{Comment}{/* }{ */}

\RestyleAlgo{ruled}

In the age of the Internet of Things, high computing performance and machine learning-based algorithms are essential. Multiple companies seek powerful and fast artificial intelligence algorithms; their goal is to increase profitability and offer convenience for users and businesses. Among the most highly sought-after algorithms are ANN algorithms. These algorithms are commonly utilized in recommendation systems [1, 2], enabling the identification of items similar to the user's favorites. This enhances user engagement with the company's product, leading to increased product usage time and greater satisfaction. The aforementioned algorithms are also applied to search engines. For instance, Elastic Search [4], the search and analytical engine powering large products such as GitHub and Netflix, is based on the HNSW algorithm [3], enabling fast delivery of optimal search results. These products can handle several million queries per second, which requires tremendous computational resources and energy. Hence, it is crucial to optimize these computations, particularly the inference of models, which is more important for users.

The RISC-V processor architecture was introduced in 2010. It is currently being developed by the community [6]. The architecture's main advantage over other popular architectures, such as x86 and ARM, is that it is open-source. Several significant chip and microcontroller manufacturers have already adopted RISC-V in their products, highlighting its benefits of openness, modularity and flexibility [7]. The modular design allows for control over the available instruction set, enabling extensions to the base ISA. A vital extension is RVV, which implements the SIMD approach - single instruction multiple data, allowing parallel processing of data in vector registers. RVV enables the efficient processing of large data arrays, resulting in considerable performance benefits for HPC (High Performance Computing) tasks [8] and artificial intelligence applications [9, 10]. Notably, the non-fixed length of RVV's vector register, determined by the VLEN parameter, simplifies development by allowing software to be written once and run on hardware with varying register lengths. Additionally, RVV permits registers to be grouped together to create a unified vector. The LMUL attribute determines the maximum number of registers that can be grouped and its optimal value for maximizing performance varies depending on the processor. The implementation of such features can simplify code writing and enhance its efficiency. 

In this study, the effectiveness of RVV in popular ANN algorithms was investigated. We implemented five ANN algorithms on RISC-V, optimized them using RVV, and experimentally compared the performance of the optimized versions of the algorithms with the reference algorithms. Furthermore, we examined the primary hotspots of the algorithms and explored the applicability of optimizations utilizing vector extensions on other architectures. In the second part of this work, we conducted a theoretical evaluation of the studied algorithms and identified the best on average vector unit configuration of the processor under certain constraints.

\section{ANN algorithms}
\label{sec:ann}
The objective of the KNN (K-Nearest Neighbors) algorithm is to locate the nearest vectors in a high-dimensional space $\mathbb{X} = \mathbb{R}^d$ of dimension $d$. The algorithm calculates the distance to the training sample objects $x \in \mathbb{X}$ for each query vector $q$. Then, the training sample objects are ordered based on their distance to $q$: $\rho(q, x_1) \le \rho(q, x_2) \le ... \le \rho(q, x_l)$, and subsequently, first k objects $x_1, ..., x_k$ are chosen. Unlike the exact KNN algorithm, which performs an exact search of all input objects to calculate distance and locate the nearest ones to the requested object, approximate search algorithms utilize heuristics to accelerate the neighbor search process. Nevertheless, this approach entails a trade-off between the running time of the algorithm and its accuracy. In highload systems, faster algorithms with minimal loss of accuracy are prioritized because the speed of the algorithm directly affects how quickly users receive results.

Several ANN algorithms, such as HNSW, FANNG, and KGraph, are based on graphs and trees [11]. Graphs enable the detection of local connections among neighbors and are more adaptable in building the index of vectors due to their extensible structure. Additionally to graph algorithms, there are algorithms of the LSH family [12]. This method involves dividing space through random hyperplanes that define a hash function with locality properties. Similar hashes identify close objects. 

In this section, we select and describe several popular ANN algorithms.

\subsection{IVFFlat}

IVFFlat [13] is one of the algorithms in the IVF (Inverted File-Based) family that utilizes a technique of dividing search space into $nprobe$
non-overlapping cells. The assumption is that objects with similar semantics are located within such a cell. Furthermore, the algorithm 
creates an inverted file, which is a data structure that maps objects and the regions to which they belong. 
Therefore, the vector space is divided using the following method. All vectors in the set $\mathbb{X}$ are clustered into $N$ cells using, 
for example, the k-means algorithm. The vectors are assigned the id of the nearest cluster, forming an inverted file index, which can be 
used to quickly find out to which region each object belongs. During the process of searching for the nearest neighbors of the query vector 
$q$, the distance $\rho$ (usually the Euclidean distance) between the $q$ and the centroids of the found clusters is calculated, then 
all vectors belonging to the nearest cluster are sorted by the distance to $q$ and $k$ nearest neighbors are selected. This method 
demonstrates fast search speed in practice with only a minor decrease in quality.

\subsection{IVFPQ}
The IVFPQ algorithm as well as IVFFlat utilizes the IVF structure to depict the connection between vectors and clusters acquired by applying the k-means to the vector space $\mathbb{X}$, but employing product quantization technique [14]. Quantization is a compression method that reduces the spatial dimensions of data to ensure efficient storage and quicker processing times. To quantize the training sample, we do not use the original data, but the component-wise difference between the vector and the radius vector of the centroid of the cluster to which the training vector belongs. This centers all vectors while also preserving the distances between them and the query vector. Next, the dataset is divided into $M$ chunks of equal length $d_{sub}$ along the vector components. Each chunk is partitioned into $k_{sub}$ clusters using the k-means algorithm. The id of the nearest cluster is assigned to each chunk of each vector. The resulting compressed data is stored in an inverted file. The $q$ vector itself is also divided into $M$ chunks, which are used to construct a distance table $dis\_table$ with dimensions $(M, k_{sub})$. The table contains the distances between each chunk of the query vector and each centroid, recorded in the corresponding cell. The calculation of the distance between the vector $q$ and the sample training vector $x_i$ can be done using the following formula: $dis(x_i, q) = \sum_{k=0..M}{dis\_table(x_i(k), k)}$.

\subsection{NSW}
A "small world" graph is a structure in which any two pairs of vertices are not connected by an edge with a high probability, but at the same time are asymptotically reachable in $O(log(N))$ steps. This is accomplished by the graph having "short" edges that connect adjacent vertices as well as opposite "long" edges that reduce the mathematical expectation of the shortest path between vertices. The authors of the NSW (Navigable Small World) algorithm [15] propose to use such a graph to solve the approximate nearest neighbor problem. The algorithm is based on an estimation of the Delaunay graph with the addition of random edges between non-adjacent vertices, and performs a greedy nearest neighbor 
search.

\subsection{HNSW}
The HNSW (Hierarchical Navigable Small World) algorithm [3] is a development of the NSW concept. HNSW extracts subgraphs from the graph and conducts iterative searches on them. The dataset is randomly divided into equal parts - $D_0, ..., D_{n - 1}$, with all the original points placed on the zero layer. On the k-th layer, the points $\bigcup\limits_{i = 0}^{n - 1 - k} D_i$ are arranged. A Small World graph is constructed on each layer, without "long edges". When searching, the process begins at the top layer $D_{n-1}$, and a greedy algorithm is used to locate the neighbor closest to the query vector. We move to the next layer when a local minimum is reached, i.e. there are no neighbors closer to the query vector in the current iteration.

\subsection{Annoy}
Annoy (Approximate Nearest Neighbors Oh Yeah) [2] algorithm constructs a set of trees by partitioning space with a hyperplane between two randomly chosen points numerous times. Generally, each resulting bin contains a small number of objects. During the neighbor search phase, the algorithm maintains a priority queue, which stores the roots of the trees at the initial time point. Subsequently, it identifies the vertex from the queue which is closest to the query point. If this vertex is a leaf, we store the sampling found. Otherwise, two new vertices are added to the queue and the process is repeated until the constructed sampling is equal to or greater than the number of searched neighbors. Then, k nearest neighbors are selected.

\section{Algorithms vectorization}
To perform the optimizations, we selected several open-source libraries that implement the algorithms listed in Section \ref{sec:ann}.
\begin{itemize}
  \item Faiss [16], developed by Facebook, is a library that offers efficient implementations of various algorithms such as IVF and HNSW 
  families. It requires external dependencies in the form of MKL, or BLAS + LAPACK, and OpenMP for parallelization of heavy single-type operations. Some algorithms support SIMD instructions using SSE3, AVX, AVX2, AVX512.
  \item Annoy [2] is Spotify's solution that is implemented in their recommendation systems. There are optimizations utilizing AVX and AVX512.
  \item NMSLIB [17] is a widely-used tool for identifying the closest neighboring objects in non-metric and generic spaces. Amazon Elasticsearch Service [18] specifically employs it. You can also build an extended library with dependencies on Boost, GNU Scientific Library, and Eigen3. The tool is additionally optimized using SSE2, AVX and Neon instructions.
\end{itemize}

Further in this section, we show which functions are most commonly vectorized in ANN algorithms.

We investigated which functions used in ANN are most often vectorized. Table~\ref{vec-table} shows which functions or structures of the five algorithms described in Section \ref{sec:ann} have vector optimizations using at least one vector extension. The table is categorised into two sections: index building and search. The plus denotes the optimized code sections that are directly or optionally utilized in the algorithm. It is worth noting that Annoy and NMSLIB only vectorize distance calculation functions between vectors for both index building and nearest vector searching. In contrast, Faiss provides a wider range of optimizations. In addition to vectorizing distance calculation functions, Faiss also vectorizes some functions of quantizers, i.e. objects that map vectors into inverted lists. Accordingly, quantizers are only used in the algorithms of the IVF family and are user-selectable as index parameters. In the HNSW implementation, only the minimax heap, which optimizes the vector search process, is vectorized. The remaining functions are utilized in either the IVFFlat and IVFPQ algorithms or other unexamined algorithms of the Faiss library. Nevertheless, distance computation is the primary bottleneck of all the considered algorithms.

\begin{table}
  \centering
  \caption{ANN Operations}
  \renewcommand{\arraystretch}{1.1}
  \label{vec-table}
  \begin{tabular}{l|l|c|c|c|c|c}
    \toprule
    \multirow{2}{*}{~} & \multirow{2}{*}{~} & \multicolumn{3}{|c|}{Faiss} & Annoy & NMSLIB \\ \cline{3-7}
    ~ & ~ & IVFFlat & IVFPQ & HNSW & Annoy & NSW \\
    \midrule
    \multirow{13}{*}{\begin{turn}{90}Build index\end{turn}} & Distance & + & + & + & + & + \\ \cline{2-7}
    ~ & L2 squared norm & + & + & ~ & ~ & ~ \\ \cline{2-7}
    ~ & Multiply-add & ~ & + & ~ & ~ & ~ \\ \cline{2-7}
    ~ & Find minimum value & + & + & ~ & ~ & ~ \\ \cline{2-7}
    ~ & Heap with buckets & ~ & ~ & ~ & ~ & ~ \\ \cline{2-7}
    ~ & Minimax heap & ~ & ~ & ~ & ~ & ~ \\ \cline{2-7}
    ~ & Distance to code & ~ & ~ & ~ & ~ & ~ \\ \cline{2-7}
    ~ & Matrix transpose & ~ & + & ~ & ~ & ~ \\ \cline{2-7}
    ~ & Partition & ~ & ~ & ~ & ~ & ~ \\ \cline{2-7}
    ~ & ProductQuantizer & + & + & ~ & ~ & ~ \\ \cline{2-7}
    ~ & LocalSearchQuantizer & + & + & ~ & ~ & ~ \\ \cline{2-7}
    ~ & ScalarQuantizer & + & + & ~ & ~ & ~ \\ \cline{2-7}
    ~ & ResidualQuantizer & + & + & ~ & ~ & ~ \\ \hline
    \multirow{13}{*}{\begin{turn}{90}Search\end{turn}} & Distance & + & + & + & + & + \\ \cline{2-7}
    ~ & L2 squared norm & + & + & ~ & ~ & ~ \\ \cline{2-7}
    ~ & Multiply-add & ~ & + & ~ & ~ & ~ \\ \cline{2-7}
    ~ & Find minimum value & + & + & ~ & ~ & ~ \\ \cline{2-7}
    ~ & Heap with buckets & ~ & ~ & ~ & ~ & ~ \\ \cline{2-7}
    ~ & Minimax heap & ~ & ~ & + & ~ & ~ \\ \cline{2-7}
    ~ & Distance to code & ~ & + & ~ & ~ & ~ \\ \cline{2-7}
    ~ & Matrix transpose & ~ & + & ~ & ~ & ~ \\ \cline{2-7}
    ~ & Partition & ~ & ~ & ~ & ~ & ~ \\ \cline{2-7}
    ~ & ProductQuantizer & + & + & ~ & ~ & ~ \\ \cline{2-7}
    ~ & LocalSearchQuantizer & + & + & ~ & ~ & ~ \\ \cline{2-7}
    ~ & ScalarQuantizer & + & + & ~ & ~ & ~ \\ \cline{2-7}
    ~ & ResidualQuantizer & + & + & ~ & ~ & ~ \\
    \bottomrule
  \end{tabular}
\end{table}

The proximity metric of vectors is often either the Euclidean distance between them or the cosine distance calculated via the scalar product. Therefore, to enhance the efficiency of algorithms, we have optimized these functions in each of our selected libraries by employing RVV 0.7.1 vector intrinsics. The pseudocode of the optimized functions is presented in Appendix A.

\section{Experimental evaluation}
\subsection{Setup}
The experiments were carried out on the Lichee Pi 4A board [21].

In our experiment, we measure the performance of index building and vector search on two datasets presented in Table ~\ref{datasets-table}. Epsilon binary classification dataset with 500,000 objects, each containing 2000 features. The pre-processed data was taken from the LIBSVM website [19]. Due to the low performance of the Lichee Pi board, we reduced the dataset to 40,000 objects by taking the first 32,000 objects from the training dataset and 8,000 objects from the test dataset. 

The experiment also uses the GloVe [20] dataset of pre-trained word embeddings built from the Wikipedia 2014 and Gigaword 5 text corpora, which has several versions with different numbers of features. For this study, we chose the version with 100 features.

\begin{table}[h]
  \caption{Datasets summary}
  \renewcommand{\arraystretch}{1.1}
  \label{datasets-table}
  \centering
  \begin{tabular}{lccc}
    \toprule
    Dataset & Train samples & Test samples & Features \\
    \midrule
    Epsilon (cropped) & 32 000 & 8000 & 2000 \\
    GloVe & 320 000 & 80 000 & 100 \\ 
    \bottomrule
  \end{tabular}
\end{table}

\subsection{Algorithm parameters}
The performance measurements were conducted with identical parameters on both datasets, except for the IVFPQ algorithm. In all algorithms, Euclidean distance was employed, and the number of neighbors in search $k$ was set to 10. From the Faiss library we used IndexIVFFFlat, IndexIVFPQ and IndexHNSWFlat implementations. The IVF family of algorithms share several common parameters and configurations:

\begin{itemize}
  \item $nlist$ regulates the number of clusters that are acquired using k-means whilst building an index. Based on the guidelines of the developers of Faiss, for datasets with less than one million vectors, we adopted a value of $nlist$ equal to $4 \sqrt{N}$, where $N$ is the size of the dataset.
  \item For vector searching, we choose $nprobe = 8$ cells to search.
  \item We also employed the IndexFlatL2 quantiser which entirely stores the vectors and carries out an exact search.
\end{itemize}

The following parameter values were selected for the algorithms individually:

\begin{itemize}
  \item For the IVFPQ method, two more parameters are used, namely the size of a chunk into which the input vector is divided, and the number of bits used to encode a component of the chunk. Both values are set to 8 for the epsilon dataset, and 4 and 8, respectively, for GloVe.
  \item HNSW algorithm takes only one argument, which is the $M$ parameter, representing the number of edges incident to the current layer's vertex.
  \item Annoy allows us to configure the number of trees to be used for index building. In our experiment, we set the value of $n\_trees$ to 64. The index is created with the AnnoyIndexSingleThreadedBuildPolicy, which disables parallelization with pthreads.
  \item The NSW algorithm from the NMSLIB library regulates the quality of construction for the small world graph with $efConstruction$ parameter and the search accuracy with the $efSearch$ parameter. They are set to 256 and 128, respectively. Furthermore, similar to the $M$ parameter in the HNSW algorithm, we set $NN=16$.
\end{itemize}

\subsection{Performance results}
\begin{figure}%[H]
  \center{\includegraphics[scale=0.3]{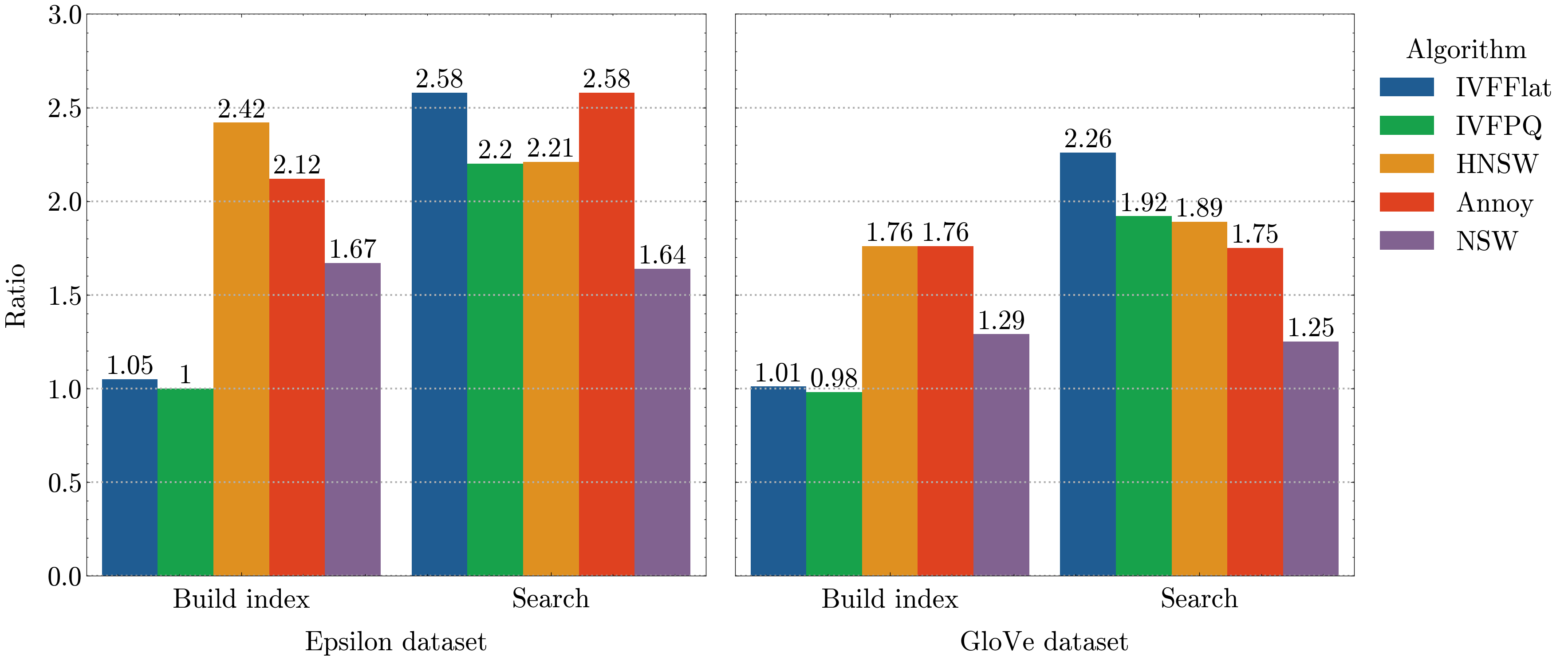}}
  \caption{ANN algorithms' acceleration.}
  \label{epsilon-perf-figure}
\end{figure}

We conducted performance tests in single-threaded mode on a single core, with the algorithm being executed ten times to extract the median time value. The tests were performed with and without RVV 0.7.1 on a Lichee Pi 4A. Figure ~\ref{epsilon-perf-figure} show the speedup of index building and nearest neighbor search iteratively on single vector for the epsilon and GloVe datasets, respectively. The speedup is calculated by normalising RVV code execution time with the time taken by the scalar code.

On the wide epsilon dataset, the performance of some algorithms increased by up to 2.58 times using RVV. The index building time in the IVFFlat and IVFPQ algorithms did not change due to the fact that the block matrix multiplication performed by the OpenBLAS library is used for clustering and distance between vectors. In contrast, the average search time was found to be more than halved in comparison to the generic version. In the case of the GloVe dataset, whose data dimensionality is 20 times lower than that of the epsilon dataset, the acceleration is lower - up to 2.26 times. The NSW algorithm exhibited the least degree of speedup, with an improvement of approximately 25\%

\section{Theoretical performance evaluation}

In order to ascertain the CPU vector unit configuration that yields the greatest average performance of the previously optimized ANN family algorithms, 
we implemented a simple parametrized vector unit model with vector instruction simulation and L2 and dot product computation. This model provides 
an upper bound on the performance of the algorithms under a number of assumptions and constraints. These include:

\begin{itemize}
  \item The data in the vector unit arrives continuously and is distributed uniformly during the operation of the vectorized prior functions. 
  The limitation is the memory controller bandwidth.
  \item The operation of the vector unit is simulated with the parameters described below. The model simulates only the previously 
  vectorized functions and the vector instructions used in them. We consider all other processor parameters and blocks to be fixed.
\end{itemize}

The model is designed to perform operations on 32-bit floating-point numbers. Additionally, the model permits the user to customize certain vector unit parameters. The five ANN algorithms were optimized using the following parameters:

\begin{itemize}
  \item $vlen$ - vector register length in bits.
  \item $k$ - number of vector registers.
  \item $n$ - number of adders.
  \item $m$ - number of fused MAC.
\end{itemize}

\begin{figure}[h]
  \center{\includegraphics[scale=0.5]{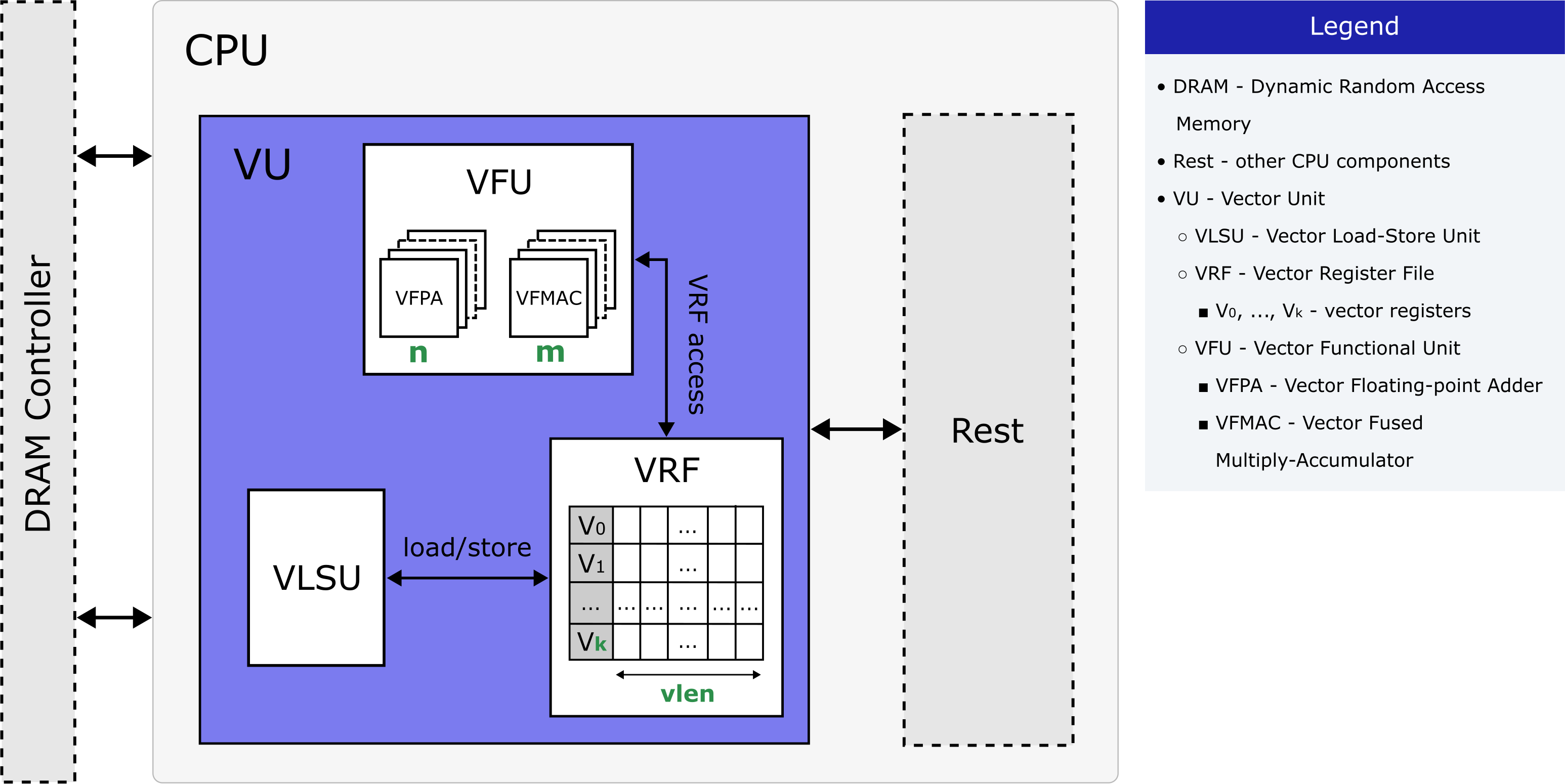}}
  \caption{Vector unit configuration.}
  \label{vector-unit-figure}
\end{figure}

This configuration of the parameterized vector block is illustrated in Figure ~\ref{vector-unit-figure}. In our configuration, separate multipliers are not employed, as it is more efficient to utilize fused MAC in ANN context.

Additionally, the following vector instruction execution times in clock cycles (hereafter CC) were utilized in the experiments:

\begin{enumerate}
  \item $add$ - addition of two vector registers (3 CC).
  \item $macc$ - multiply-addition of two vector registers, followed by the accumulation of the result in the third vector register (6 CC).
  \item $load$ - loading data into vector registers (1 CC).
  \item $setvl$ - vector length setting (1 CC).
  \item $mv\_v\_to\_s$ - storing the first value of the vector register into a scalar register (1 CC).
  \item $mv\_s\_to\_v$ - duplicate a floating-point scalar operand to a vector register (1 CC).
\end{enumerate}

Experiments were conducted with vectors of 2000 features. The CPU frequency was fixed at 1.85 GHz, and the memory controller bandwidth was set to 29.8 GB/s, as in the Lichee Pi 4A [21]. The theoretical performance estimates of the ANN family algorithms achievable on this model are presented in Table ~\ref{perf-table}. The key result is the best on average configuration. The key result is the best average configuration defined as the one that maximizes the average sum of the number of queries per second (Q/s) of each algorithm.

\begin{table}[H]
  \caption{Best parametrized vector unit configurations for ANN algorithms}
  \renewcommand{\arraystretch}{1.1}
  \label{perf-table}
  \centering
  \begin{tabular}{c|c|c|c|c|c|c}
    \toprule
    ~ & IVFFlat & IVFPQ & HNSW & Annoy & NSW & Best on average \\
    \midrule
    k & 4 & 4 & 4 & 4 & 4 & 4 \\ \hline
    vlen & 2048 & 512 & 2048 & 16384 & 2048 & 512 \\ \hline
    n & 8 & 16 & 8 & 512 & 8 & 16 \\ \hline
    m & 64 & 16 & 64 & 8 & 64 & 16 \\ \hline
    bandwidth (GB/s) & 29.5 & 27.3 & 29.5 & 25.4 & 29.5 & 24.2\tablefootnote{mean value by all the algorithms with this configuration.} \\ \hline
    Q/s & 367 & 3331 & 4520 & 1218 & 459 & 1666 \\
    \bottomrule
  \end{tabular}
\end{table}

\section{Conclusions}
In this research, we have analyzed a range of ANN algorithms and optimized the key parts of the code to show that SIMD instructions of RVV vector extension are effective in these scenarios. Our findings indicate that this approach is effective in ANN algorithms, with significant performance improvements observed in index building and nearest neighbor searches due to the vectorization of distance computation functions. These results suggest that the use of vector instructions may be beneficial in other ANN algorithms. Other functions can also be vectorized, which is likely to further improve the speed of the algorithms in addition to the distance function. Additionally, we developed a simple vector unit model for the purpose of analyzing the performance of a set of ANN algorithms under specific tunable parameters. This analysis allowed us to determine the best on average configuration for these algorithms.

\clearpage

\setcounter{table}{0}

\renewcommand{\thetable}{A\arabic{algorithm}}

\appendix
\section*{Appendix A. Vectorized algorithms}
\label{sec:appendix_a}

\begin{algorithm}%[hbt!]
  \caption{Vectorized L2 distance}\label{alg:l2_distance}
  \KwData{$N, a, b$}
  \KwResult{$\lVert{a - b}\rVert$}
    $vl \gets \text{vsetvl}(N)$\;
    $s_{v} \gets \text{vfmv.v.f}(0, vl)$\;
    \For{$i \gets 0$ \KwTo $N$ \KwBy $vl$} {
      $a_{v} \gets \text{vle32.v}(a + i, vl)$\;
      $b_{v} \gets \text{vle32.v}(b + i, vl)$\;
      $d_{v} \gets \text{vfsub.vv}(a_{v}, b_{v}, vl)$\;
      $s_{v} \gets \text{vfmacc.vv}(s_{v}, d_{v}, d_{v}, vl)$\;
    }
    $dis_{v} \gets \text{vfredosum.vs}(s_{v}, vl)$\;
  
    /* Handle leftovers */
  
    \Return{dis}
\end{algorithm}

\begin{algorithm}
  \caption{Vectorized dot product}\label{alg:dot_product}
  \KwData{$N, a, b$}
  \KwResult{$\lVert{a - b}\rVert$}
    $vl \gets \text{vsetvl}(N)$\;
    $s_{v} \gets \text{vfmv.v.f}(0, vl)$\;
    \For{$i \gets 0$ \KwTo $N$ \KwBy $vl$} {
      $a_{v} \gets \text{vle32.v}(a + i, vl)$\;
      $b_{v} \gets \text{vle32.v}(b + i, vl)$\;
      $s_{v} \gets \text{vfmacc.vv}(s_{v}, a_{v}, b_{v}, vl)$\;
    }
    $dis_{v} \gets \text{vfredosum.vs}(s_{v}, vl)$\;
  
    /* Handle leftovers */
  
    \Return{dis}
\end{algorithm}

\end{document}